\documentclass{article}
\usepackage{ijcai16}

\usepackage{times}

\usepackage{graphicx}
\usepackage{amsmath,bm}
\usepackage{amsfonts}
\usepackage{mathrsfs}
\usepackage{amsbsy}
\usepackage{paralist}
\usepackage{multirow}
\usepackage{tabularx,color}
\usepackage[ruled,vlined]{algorithm2e}

\usepackage{fancyhdr} 

\usepackage{CJK}
\usepackage{hhline}

\newcommand\Tstrut{\rule{0pt}{2.4ex}}         

\newlength{\Oldarrayrulewidth}
\newcommand{\Cline}[2]{%
  \noalign{\global\setlength{\Oldarrayrulewidth}{\arrayrulewidth}}%
  \noalign{\global\setlength{\arrayrulewidth}{#1}}\cline{#2}%
  \noalign{\global\setlength{\arrayrulewidth}{\Oldarrayrulewidth}}}
\DeclareMathOperator{\Sim}{sim}
\newcolumntype{Y}{>{\centering\arraybackslash}X}

\newcommand{\name}{Bi-PageRank-HITS{}}

\title{StalemateBreaker: A Proactive Content-Introducing Approach\\ to Automatic Human-Computer Conversation}

\author{Xiang Li,$^{1,}$\thanks{Contribution partially done during internship at Baidu Inc. 
Corresponding author: Rui Yan (yanrui02@baidu.com).} Lili Mou,$^{1,2}$ Rui Yan,$^3$ Ming Zhang$^1$ \\
$^1$School of EECS, Peking University, China\quad {\tt \{lixiang.eecs,mzhang\_cs\}@pku.edu.cn} \\
$^2$Key Laboratory of High Confidence Software Technologies (Peking University),\\
Ministry of Education, China\quad {\tt doublepower.mou@gmail.com}\\
$^3$Natural Language Processing Department, Baidu Inc., China\quad {\tt yanrui02@baidu.com}
}

\begin{document}
\begin{CJK*}{UTF8}{gkai}
\maketitle

\renewcommand{\headrulewidth}{0pt}
\cfoot{}
\chead{Accepted by the 25th International Joint Conference on Artificial Intelligence (IJCAI 2016)}
\thispagestyle{fancy}

\begin{abstract}
 Existing open-domain human-computer conversation systems are typically \textit{passive}: they either synthesize or retrieve a reply provided a human-issued utterance. It is generally presumed that humans should take the role to lead the conversation and introduce new content when a stalemate occurs, and that the computer only needs to ``respond.'' In this paper, we propose \textsc{StalemateBreaker}, a conversation system that can proactively introduce new content when appropriate. We design a pipeline to determine when, what, and how to introduce new content during human-computer conversation. We further propose a novel reranking algorithm \name\ to enable rich interaction between conversation context and candidate replies. Experiments show that both the content-introducing approach and the reranking algorithm are effective. Our full \textsc{StalemateBreaker} model outperforms a state-of-the-practice conversation system by $+14.4\%$ p@1 when a stalemate occurs.
\end{abstract}

\section{Introduction}\label{sec:introduction} 
Automatic human-computer conversation is believed to be one of the most challenging problems in artificial intelligence (AI). For decades, researchers have developed various systems based on human-crafted rules \cite{webb2000rule,varges2009leveraging}, information retrieval methods \cite{misu2007speech,sigir}, or natural language generators like neural networks \cite{shang2015neural}. In these systems, the computer either searches or synthesizes a \textit{reply} given an utterance (called \textit{query}) issued by a user. It is generally presumed that ``humans'' should play a leading role in human-computer conversation. Hence traditional AI conversation is a passive process: what a computer does is just to ``respond.''

\begin{figure}[!t]
\centering
\includegraphics[width=.3\textwidth]{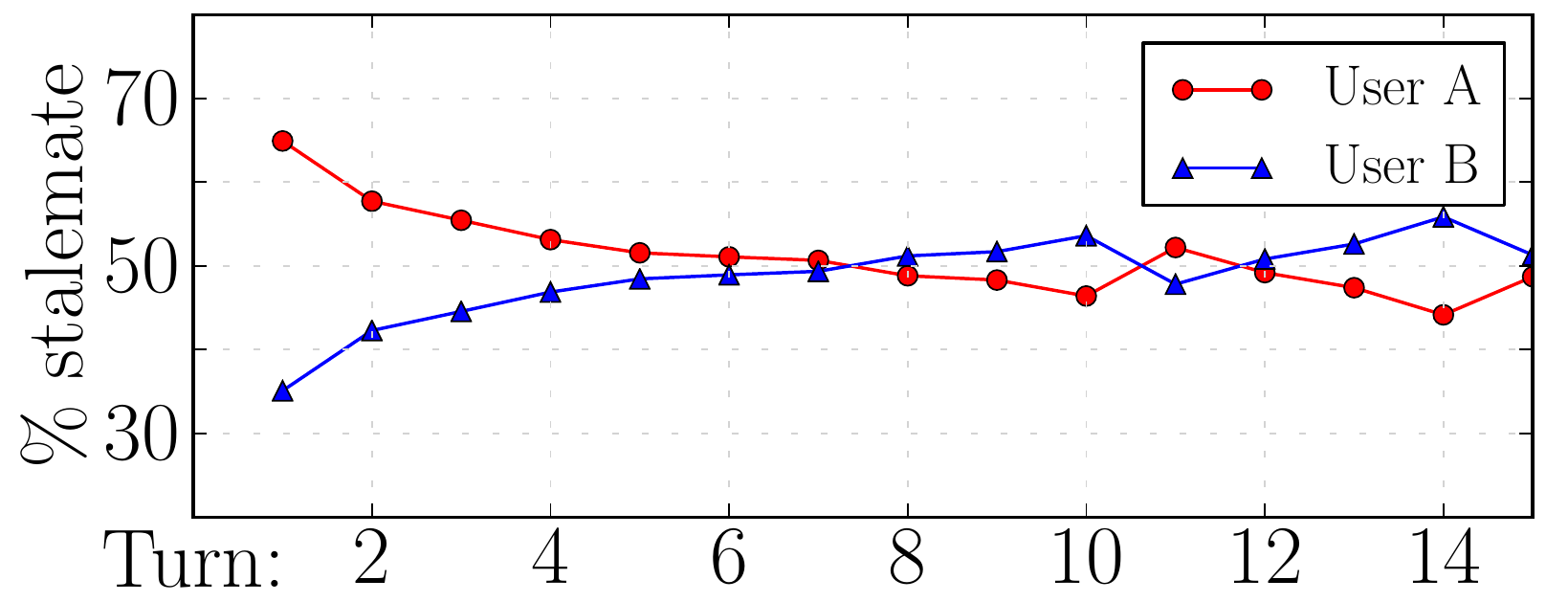}
\caption{In a multi-turn human-human conversation started by User A, we plot the percentage at which each user should take the role of conversation leading. In other words, we detect how likely a stalemate occurs to a particular user.}\label{fig:stat}
\end{figure}

In human-human conversation, however, both participants have the duty to play a leading role in a continuous dialogue session. The phenomenon is supported by the statistics of conversation data collected from an online forum\footnote{{http://www.douban.com}} (Figure~\ref{fig:stat}). Our observation is that, shortly after the conversation begins, both parts are likely to be the stalemate breaker. If only one side keeps finding something to talk while the other side responds in an unmindful way, the conversation becomes less attractive and is likely to end pretty soon. Therefore, in human-computer conversation, the computer side should also be initiative and introduce new content when necessary.

The problem of content introducing is also raised from industry. Although real-world conversation will end sooner or later, industrial conversation systems shall always try to attract users for commercial purposes (except when users explicitly terminate a session). Thus, stalemate breaking is of particular importance to industrial conversation products.

Existing mixed-initiative dialogue systems are typically designed in vertical domains. For example, \citeauthor{a53}~\shortcite{a53} develop a rule-based system, named TRAINS-95, in the transportation domain; \citeauthor{transition}~\shortcite{transition} leverage  pre-defined topics in a museum-guiding system. Such design methodology, however, hardly applies to non-task-specific, chat-style dialogues. Since users are free to say anything, it is virtually impossible to specify rules or design templates for open-domain conversations. Moreover, the content to be introduced is nearly certain in those task-~or goal-oriented applications \cite{goal}, whereas the nature of open-domain conversations shows that a variety of replies are plausible, but some are more meaningful, and others are not. Consequently, open-domain conversations are different from task-specific dialogues; the same thing holds for content introducing in these two scenarios.

In this paper, we propose \textsc{StalemateBreaker}, a conversation system that can proactively introduce new content during human-computer conversations. We first detect whether a stalemate occurs by keyword filtering like ``\dots'' or ``Errr,'' so that our system knows \textit{when} the stalemate-breaking mechanism should be triggered. To determine \textit{what} to introduce, we backtrack previous utterances~(called \textit{context}) in a dialogue session and apply named entity recognition in the context. The detected entities are searched for more related entities in a large knowledge graph. All these entities are used to retrieve candidate replies in a conversation database. We believe named entities highly reflect users' interest and provide informative clues for content introducing.  We then propose a \name\ algorithm to address \textit{how} to introduce. By matching the relationship between the context and candidate replies (containing the entities) in a reinforced co-ranking manner, we obtain a ranked list, indicating the relevance of each candidate reply. In this way, our system is well aware of \textit{when}, \textit{what}, and \textit{how} to proactively introduce new context in a continuous human-computer conversation.

We build our proactive content-introducing system upon a large conversation database for retrieval (9.8 million candidate query-reply pairs) plus an external knowledge graph~(3.7 million tuples). We evaluate our system on conversation logs from real-world users. Our approach outperforms several strong baselines as well as a state-of-the-practice system. 


\section{Related Work}


\subsection{Dialogue systems}
\noindent\quad$\bullet$ \textbf{Domain-specific systems.} Automatic human-computer conversation has long attracted attention in several vertical domains~\cite{bernsen1994dedicated,rickel2000task}. In such task- or goal-oriented applications, researchers have developed mixed-initiative systems to more effectively guide users in conversation. Several examples include TRAINS-95 for route planning \cite{a53}, MIMIC for movie show-time information \cite{a52}, and AutoTutor for learner advising \cite{graesser2005autotutor}.
These systems rely heavily on human-designed rules or templates.  Other systems may require intensive domain knowledge to be initiative in conversation, e.g., museum guiders \cite{transition}, children companion systems \cite{adam2010flexible,macias2012coherent}, etc.

\noindent\quad$\bullet$ \textbf{Open-domain systems.} Human-engineered rules may also be applied to the open domain as \citeauthor{sigdial}~\shortcite{sigdial} do, but their generated sentences are subject to 7 predefined forms and hence are highly restricted; they leverage external knowledge bases to enhance content in the responses. Recently, more and more studies and systems are tackling the real challenges of the open domain:  great flexibility and diversity. Retrieval-based methods ``query'' a user-issued utterance in a large database of existing dialogues, and return appropriate responses \cite{higashinaka2014syntactic,ji2014information}. Generative methods---typically using statistical machine translation techinques \cite{ritter2011data,sugiyama2013open,mairesse2014stochastic} or neural networks \cite{shang2015neural,sordoni2015neural}---can synthesize new replies, although the generated sentences are not guaranteed to be a legitimate natural language text. Industrial products like {\tt Siri} of Apple, {\tt Xiaobing} of Microsoft, and {\tt Xiaodu} of Baidu, are among state-of-the-practice systems; they are increasingly affecting people's everyday life.

To the best of our knowledge, existing open-domain chatbot-like conversation are a \textit{passive} process: the computer only needs to ``respond'' to human inputs and does not take the role of conversation leading. Instead, we propose a \textit{proactive} system, which can determine  when, what, and how to be proactive and to introduce new content into the conversation.

\subsection{Random Walk-Based Ranking}
Our system follows a retrieval-and-reranking schema to select replies from a candidate pool. In this part, we briefly review (re)ranking algorithms like PageRank and its variants.

In the field of information retrieval, research shows that random walks over hyper-link graphs, i.e., PageRank, can reflect the relationship between different web pages and rank high important ones \cite{page1999pagerank}. Many studies are devoted to the application and extention of PageRank  \cite{haveliwala2002topic,jeh2003scaling}. Random walks over bipartite graphs can model two heterogeneous types of items, e.g., the HITS algorithm for queries and documents in a click-though graph \cite{kleinberg1999authoritative,deng2009generalized}. Its variants have been widely used for ranking tasks in the information retrieval community \cite{cao2008context,song2012query}. In our scenario, the matching between user utterances and candidate replies can also be modeled as a bipartite graph; to enhance their interaction, we extend existing models and propose \name, which is a novel algorithm for reranking.

\begin{figure*}[t]
 \centering
 \includegraphics[width=.82\textwidth]{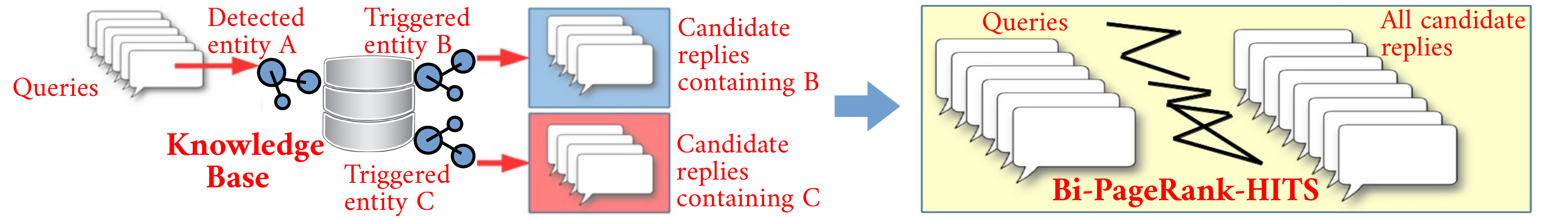}
 \caption{Overview of our \textsc{StalemateBreaker} system.}
 \label{fig:architecture}
\end{figure*}

\section{The Proposed Approach}
\subsection{Architecture}

\begin{figure}
\centering
\includegraphics[width=.38\textwidth]{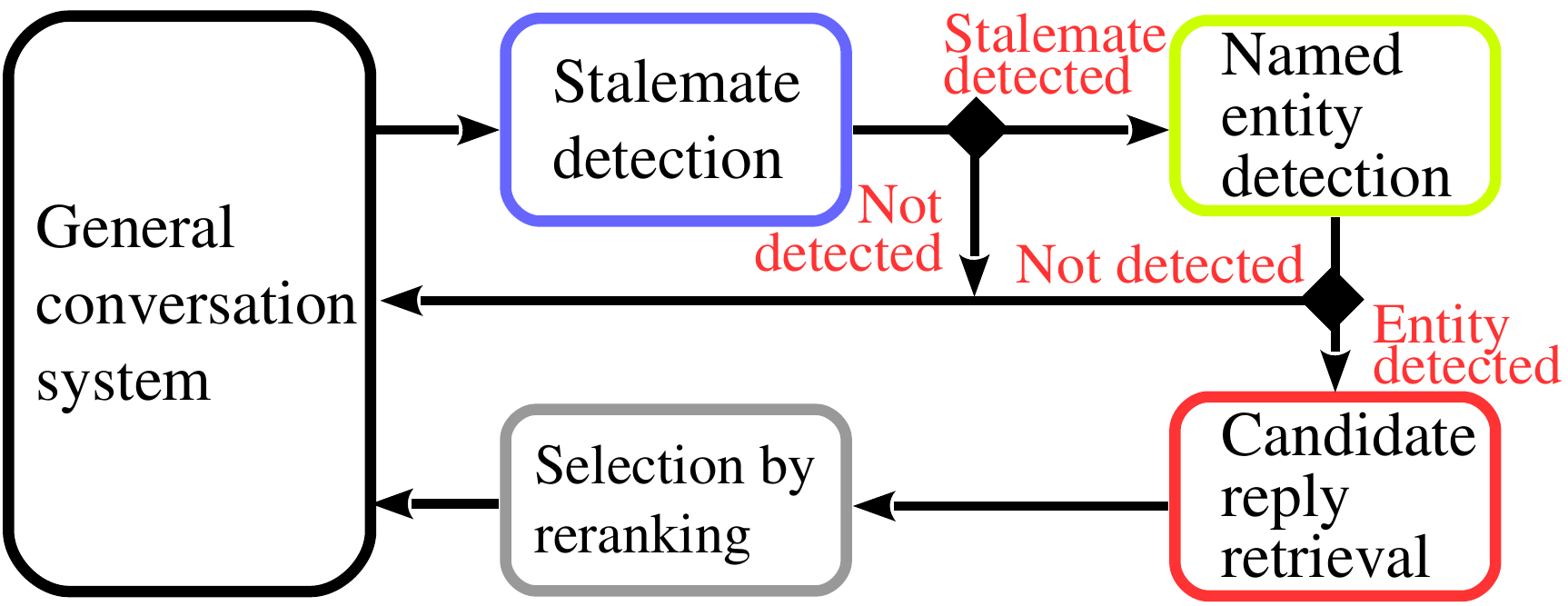}
\caption{Process flow of triggering content introducing.}\label{fig:process}
\end{figure}

Figure~\ref{fig:architecture} shows the overall architecture of our \textsc{StalemateBreaker} system; Figure~\ref{fig:process} further illustrates the process flow of content introducing. Our system comprises mainly four steps:

\textbf{Stalemate detection.} The system is built upon a conventional retrieval-based conversation system, which is typically passive. The proactive content introducing starts from stalemate detection. We apply keyword matching of meaningless expressions like ``\dots'' or ``Errr.'' In total, we have nearly a hundred filters. Although simple, the approach works in a pragmatic way and is not the main focus of this paper. In future work, we would like to apply learning-based sentence modeling (e.g., \citeauthor{sentencemodel}~\shortcite{sentencemodel}) for stalemate detection.

\textbf{Named entity detection.} Once the content-introducing mechanism is triggered, our system backtracks previous utterances in the current conversation session, and detects all named entities within a window. Currently, we keep four utterances, i.e., two turns, as context. We believe recently mentioned named entities highly reflect users' interest; hence we search related entities in a knowledge base for introducing new content. Concretely, the knowledge base is composed of tuples like $\langle e_1, e_2, w\rangle$, indicating the entity $e_1$ is correlated to $e_2$ with a weight of $w$. Notice that the relation between $e_1$ and $e_2$ in this tuple is directed because weights are asymmetric. For each entity in the current context, we search it in the knowledge base and keep top (highest weighted) five returned entities for further processing. 

In our work, the knowledge base we used was constructed from query logs of an information retrieval system. Without loss of generality, we can also exploit similar resources, e.g., ontologies \cite{ontology} or information networks \cite{information}. Leveraging knowledge bases for content introducing, in fact, is also applied in \citeauthor{sigdial} \shortcite{sigdial}. However, they plug related entities to several predefined templates for response generation, whereas we have further developed complicated retrieval-and-reranking approaches.

It should be mentioned that if the conversation is not in a stalemate or we could not recognize any named entity in previous several utterances, our system will return to the general conversation mode. In other words, the proactive content-introducing method can be viewed as an ``add-on'' mechanism to a mature conversation system.

\textbf{Candidate reply retrieval.} We then use the entities and conversation context to retrieve up to fifty candidate replies from a large pool of collected dialogue data. A candidate reply contains at least one entity. This process is accomplished by standard keyword-based retrieval methods, similar to the {\tt Lucene} system.

\textbf{Selection by reranking.} Finally, the candidate replies are reranked by a random walk-like algorithm. To enhance interaction between conversation utterances and candidate replies, we further propose \name, a novel algorithm that combines PageRank and HITS into a single framework. (See next subsection.) The highest (re)ranked candidate is selected as the reply. 


\subsection{Reranking Algorithm}\label{ss:rank}

In this part, we describe in detail the proposed \name\ algorithm, which is a combination of PageRank \cite{page1999pagerank} and HITS \cite{kleinberg1999authoritative,deng2009generalized}.\footnote{``HITS'' is the acronym of \textit{hyperlink-induced topic search}.}

We formulate the utterances in context (called \textit{queries}) and candidate replies as a bipartite graph. Following the notations in HITS, we denote queries as ``hubs'' and replies as ``authorities.'' Then our random walk-style algorithm alternately ranks either side in the query-candidate graph (PageRank step) and interacts between the two sides (HITS step).

Our intuition is that a high hub score indicates the query is important, providing a clue for content introducing; the interaction between two sides assesses the appropriateness of a reply to all queries (reweighted by query importance). Then another PageRank over authorities suggests high-quality replies; such information is propagated back to queries in a similar way. So on and so forth, the hub and authority scores are iteratively computed in a reinforced fashion. After convergence, we obtain an overall ranking list for candidate replies. Because PageRank applies to both hubs and authorities, the model in this paper extends our previous work \cite{cikm,ijcai}, and we name the algorithm \name. In the rest of this subsection, we present individual PageRank and HITS steps and then describe how they are combined.

\textbf{PageRank step.}
For either side (e.g., queries) of the query-reply bipartite graph,  we use PageRank \cite{page1999pagerank} for scoring. We do not consider the other side (e.g., replies) in this step.

Considering the set of all queries (utterances in the conversation context), we define a random walk over an undirected graph $G_q=(V_q,E_q)$, whose nodes are the queries and edges are the relationships between a query-query pair. The weight of an edge $i\rightarrow j$ is defined to be the ``similarity'' between queries $i$ and $j$, i.e.,
$M_{q,ij} =  \Sim(q_i,q_j)$, where we use the cosine measure as similarity based on two queries' \textit{tf$\cdot$idf} vectors.

To incorporate a prior distribution $\mathbf x$ over queries, we follow \citeauthor{a23}~\shortcite{a23} and define the PageRank formula as
\begin{equation}
\mathbf{q}^{(i+1)}=(1-\mu)[\text{Diag}(\mathbf{x})\text{M}_q^\top]\mathbf{q}^{(i)}+\mu \mathbf{x}\label{eqn:pagerank1}
\end{equation}

Likewise, for the set of candidate replies, we have
\begin{equation}
\mathbf r^{(i+1)}=(1-\mu)[\text{Diag}(\mathbf{y})\text{M}_r^\top]\mathbf r^{(i)}+\mu \mathbf{y}\label{eqn:pagerank2}
\end{equation}
where $[\cdot]$ denotes column normalization, that is to say, each column is the transition probability of its corresponding node. Superscripts indicate the number of iterations in a particular PageRank step (called \textit{local iteration}).

PageRank is inspired by the following intuition. A candidate reply is important if it is ``voted'' by many other candidates. The prior ($\mathbf x$ for queries or $\mathbf y$ for replies) is initialized as a uniform distribution, but may change to emphasize particular nodes suggested by the HITS step during interaction between queries and replies.



\textbf{HITS step.}
After the above step, we obtain PageRank scores for either queries or candidate replies. To propagate such information to the other side in the query-reply bipartite graph, we perform another random walk with links between queries and replies representing the structural information of hubs and authorities. 

Formally, the bipartite graph $G=(V,E)$ has vertexes  $V=\{V_q \cup V_r\}$, where $V_q$ are queries and $V_r$ are replies. We define the weight matrix by a relevance scoring function $\phi(q,r)$ between queries and replies. $\phi(\cdot,\cdot)$ was learned via a \textit{learning-to-rank} model similar to \citeauthor{burges2005learning}~\shortcite{burges2005learning} with rich features including textual similarity, translation models, as well as {\tt word2vec} word embeddings \cite{mikolov2013distributed}. In other words, $\phi(\cdot,\cdot)$ returns the relatedness between a query and a reply in the range $(0,1)$.

Moreover, since we would like to make use of PageRank scores for queries or replies, the HITS links are judged by not only the static relevance score $\phi(\cdot,\cdot)$, but also the information given by PageRank. To be concrete, the (unnormalized) weight matrix is given by either
\begin{equation}
\text{\quad \quad\quad\quad\!}\tilde{\mathbf W}_{ij} = \phi(q_i, r_j)\cdot q_i\quad\quad\quad\text{(query$\rightarrow$reply)}
\label{eqn:w1}
\end{equation}
\begin{equation}
\text{or\quad\quad\quad} \tilde{\mathbf W}_{ij} = \phi(r_i, q_j)\cdot {r}_i\quad\quad\quad\text{(reply$\rightarrow$query)}
\label{eqn:w2}
\end{equation}
Here, $q_i$ and $r_i$ are the \textit{i}-th element in the vectors $\mathbf q$ and $\mathbf r$, which are obtained in the PageRank phase by Equations~\ref{eqn:pagerank1}--\ref{eqn:pagerank2}. If information is propagated from queries to replies, we use the former equation for weight update, and \textit{vice versa}.

The mutual-reinforcing relationship of hub and authority scores can be expressed in matrix representation as follows.
\begin{align}
\mathbf{x}^{(i+1)} &= \alpha_x\cdot \big[\tilde{\mathbf W}\big]\,\mathbf{y}^{(i)}+(1-\alpha_x)\cdot \hat{\mathbf{x}}\label{eqn:HITS2}\\
\mathbf{y}^{(i+1)} &= \alpha_y\cdot \left[\big[\tilde{\mathbf W}\big]^\top\right]\mathbf{x}^{(i)}+(1-\alpha_y)\cdot\hat{\mathbf{y}} \label{eqn:HITS1}
\end{align}
where $\mathbf x$ is query scores, $\mathbf y$ reply scores; superscripts denote the number of local iterations in HITS update.
Each column of the weights is normalized to be a valid probability. Moreover, to compute the weight for $\mathbf{y}$, the matrix $\mathbf{\tilde W}^\top$ should be first row-normalized (given by $[\mathbf{\tilde W}]^\top$); otherwise, the effect of the prior $q_i$ in Equation~\ref{eqn:w1} is ruled out by column normalization.
Notice that the transition matrix is fixed in one step of HITS, but they may also change as our algorithm proceeds like the PageRank step.

In the above equations, the first term is the standard HITS, which is entirely determined by the linkage structure between hubs and authorities. The second term indicates that, with a certain probability, the hub and authority scores will be influenced by their prior scores $\hat{\mathbf x}$ and $\hat{\mathbf y}$. Such idea of combining additional information of hubs and authorities is proposed in \citeauthor{deng2009generalized}~\shortcite{deng2009generalized} and called Co-HITS.\footnote{Precisely, our model uses the Co-HITS variant. But for simplicity, we denote it as HITS, if not confused, for notational purposes.}

We define the prior score of a query to be proportional to the averaged relevance (textual similarity) score to all replies, i.e., $\hat x_i\propto\frac1{\#r'}\sum_{r'}\Sim(q_i,r')$. Likewise,
$\hat y_i\propto\frac1{\#q'}\sum_{q'}\Sim(q',r_i)$. The vectors $\hat{\mathbf x}$ and $\hat{\mathbf y}$ are also self-normalized so as to be valid probabilities.
\begin{algorithm}[!t]
\footnotesize
\label{alg}
\caption{\name}
\textbf{Input}: Queries (utterances) and candidate replies\\
\textbf{Output}: Ranking list of replies for content introducing\\
\Begin{
//Global iteration in \name\\
\Repeat{Global convergent}{
  	Update query priors\\
  	\Repeat{Local convergent}{
  		PageRank update over queries by Eqn.~\ref{eqn:pagerank1}
  	}
  	\medskip
  	Update HITS weights by Eqn.~\ref{eqn:w1} (query$\rightarrow$reply)\\
  	\Repeat{Local convergent}{
  		HITS update over query-reply bipartite graph
  	}
  	\medskip
  	Update reply priors\\
  	\Repeat{Local convergent}{
  		PageRank update over replies by Eqn.~\ref{eqn:pagerank2}
  	}
  	\medskip
  	Update HITS weights by Eqn.~\ref{eqn:w2} (reply$\rightarrow$query)\\
  	\Repeat{Local convergent}{
  		HITS update over query-reply bipartite graph
  	}
  }
\includegraphics[width=.42\textwidth]{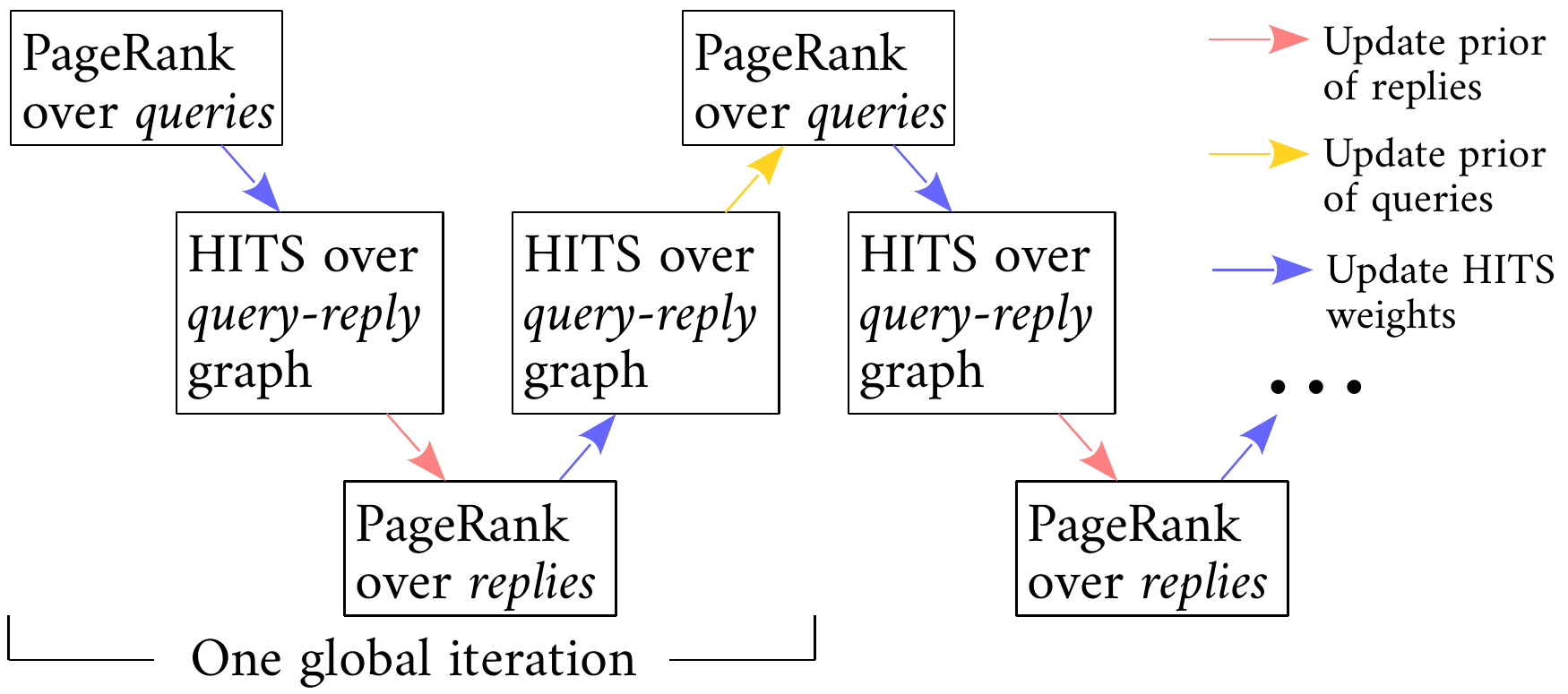}
}
\end{algorithm}

\textbf{Iteration over PageRank and HITS.}
After performing PageRank over one side (e.g., queries) of the bipartite query-reply graph and propagating the information with HITS, we shall perform another PageRank on the other side (e.g., replies). As mentioned, we use the results $\mathbf x$ or $\mathbf y$ obtained by HITS as the prior information, and recompute the transition weights in PageRank (Equation~\ref{eqn:pagerank1}~or~\ref{eqn:pagerank2}). Then, the information is propagated back by HITS, and PageRank is applied again for a better estimation.

As depicted in Algorithm~\ref{alg}, our \name\ algorithm performs PageRank and HITS steps alternately and computes the importance of a query/reply in a reinforced way. Note that the transition matrix in each step of PageRank or HITS is fixed, but they change dynamically during global iterations. Thus our model is  different from the original PageRank or HITS.

For convergence concerns, we have normalized all prior distributions and all columns in transition matrices. Therefore, the convergence of PageRank and HITS is guaranteed. For alternation between the two steps, we shall empirically analyze its convergence in Section~\ref{ss:Analysis}. In practice, we terminate our algorithm when the mean square difference between two successive HITS scores in global iteration is less than a threshold~($10^{-6}$ in our study).

After convergence, we use replies' HITS scores $\mathbf y$ for final reply selection. Compared with PageRank scores $\mathbf r$, HITS scores convey more structural information between queries and replies. A higher score indicates the reply is more appropriate for content introducing.

\section{Evaluation}
\subsection{Datasets and Experimental Setups}
Our content-introducing open-domain conversation system is built upon a large database of conversation data for retrieval. We collected massive resources from (Chinese) forums, microblog websites, and community question-answering platforms including Baidu Zhidao, Baidu Tieba, Douban forum, Sina Weibo, etc.\footnote{{{http://\{zhidao.baidu$|$tieba.baidu$|$douban$|$weibo\}.com}}}  In total, we extracted nearly 10 million query-reply pairs. Besides, we leveraged a knowledge graph mined from Baidu search logs. 


To evaluate the proposed \textsc{StalemateBreaker}, we resorted to human evaluation, following \citeauthor{ritter2011data}~\shortcite{ritter2011data} and \citeauthor{shang2015neural}~\shortcite{shang2015neural}. Objective scores like BLEU and traditional evaluation for dialogue systems (e.g., accuracy of template classification) are less applicable to our scenario, because open-domain conversation is highly diverse---one query can have a lot of suitable replies that appear different to each other. Human evaluation, on the other hand, conforms to the ultimate goal of open-domain conversation systems.
In our experiments, we used 180 sessions from real-world user conversation logs. For each entity in the context (4 previous utterances in the session), we sought 5 most related entities in the knowledge graph, and for each related entity, we retrieved top-10 candidate replies (containing the entity). 

We hired workers on a Chinese crowdsourcing platform to annotate all retrieved results with \textbf{1 Point} (appropriate) or \textbf{0 Point} (inappropriate). A candidate reply was annotated by 3 workers in an independent and blind fashion. We regarded the majority voting as the ``ground truth'' indicating whether the reply is appropriate for content introducing.  We also evaluated the kappa score: $\kappa=0.768$, showing high inner-annotator agreement \cite{fleiss1971measuring}.

Our website\footnote{https://sites.google.com/site/stalematebreaker/} provides additional data statistics and rating criterion.

\subsection{Competing Methods}

We compared the proposed proactive content-introducing method with passive conversation systems. As our work is an ``add-on'' mechanism to a deployed system, we presumed that a stalemate had been detected in our evaluation. In other cases, our proposed method does not corrupt the existing system.

For fairness, our baselines (passive conversation systems) were also aware of context information for candidate reply retrieval, since an utterance like ``Errr'' itself contains little substance. We also performed data cleaning like \citeauthor{a21}~\shortcite{a21} by removing candidates of low linguistic quality such as extremely short ones or meaningless babblings.

Regarding ranking algorithms, we compared the proposed \name\  with the following methods:
\begin{compactitem}
\item \textbf{Textual similarity.} This method ranks candidate replies according to textual similarity, predicted by a regression model with human engineered features. It is a state-of-the-practice system\footnote{http://duer.baidu.com} which our experimental environment was built upon.

\item \textbf{Reply PageRank.} PageRank is a widely used ranking algorithm \cite{page1999pagerank}, and is, actually, a component of the \name\ model.

\item \textbf{HITS.} HITS is a link analysis algorithm suitable for modeling bipartite graphs like web click-through data \cite{kleinberg1999authoritative}.

\item \textbf{Co-HITS.} Co-HITS \cite{deng2009generalized}, a variant of HITS, is another component of our \name\ model. Details are described in Section~\ref{ss:rank}.

\end{compactitem}

\begin{table}[t]
\centering
\small
\resizebox{.47\textwidth}{!}{
\begin{tabularx}{.52\textwidth}{|p{.09\textwidth}|l|Y|*{2}{|Y}|}
\hline
Group & Reranking Method & p@1& MAP & nDCG \\ \hline
\multirow{5}{1.5cm}{No content introducing}
&Textural similarity$^\dag$ &0.406  &0.498  &0.648 \\
&HITS & 0.467 & 0.550 & 0.684 \\
&Reply PageRank & 0.428 & 0.514 & 0.660 \\
&Co-HITS & 0.472 & 0.552 & 0.686 \\
&\name & \textit{0.483} & \textbf{0.556} & \textbf{0.690} \\ \hline  \Cline{1.6pt}{4-5}
\Tstrut \multirow{5}{1.7cm}{Entity-based content introducing}\!\!\!
&Textural similarity & 0.511  & 0.551  & 0.742 \\
&HITS & 0.494 & 0.542 & 0.733 \\
&Reply PageRank & 0.467 & 0.436 & 0.660 \\
&Co-HITS & 0.511 & 0.555 & 0.743 \\
&\name$^\ddag$ & \textbf{0.550} & \textbf{0.562} & \textbf{0.750} \\ \hline
\end{tabularx}
}
\caption{Performance of our method and baseline systems. $^\dag$A state-of-the-practice system which our model is built upon. $^\ddag$The full \textsc{StalemateBreaker} system. Notice that the MAP and nDCG metrics are not comparable outside a group because the retrieved candidates are different.}
\label{tab:result}
\end{table}



\subsection{Overall Performance}
We first evaluated the performance using the p@1 metric, which is believed to be the most direct judgment of conversation systems \cite{wang2013dataset,shang2015neural}. It reflects exactly the ``accuracy'' of the highest-ranked reply. Further, we applied mean average precision (MAP) and normalized discounted cumulative gain~(nDCG), as both our system and baselines return a ranking list containing multiple candidates.
For details of our metrics, we refer interested readers to \citeauthor{jarvelin2002cumulated}~\shortcite{jarvelin2002cumulated} and \citeauthor{kishida2005property}~\shortcite{kishida2005property}.
Formulas are also listed on our website.

Table~\ref{tab:result} shows the performance of our \textsc{StalemateBreaker} system as well as a variety of baselines.

The main result is that proactive entity-based content introducing is generally better (higher p@1) than passive conversation systems regardless of the ranking algorithm. Although passive systems can reply to important previous utterances in the current conversation session to some extent (because the baselines are also context-aware), they are more likely to repeat existing topics and be stuck in the stalemate.

By contrast, our \textsc{StalemateBreaker} proactively seees a knowledge base, retrieves candidate replies that contain related entities, and reranks more appropriate ones. Thus, entity-based content introducing methods yield higher scores in terms of p@1, i.e., accuracy of the top-ranked reply.

Regarding the \name\ ranking algorithm, it outperforms its base models, PageRank and HITS/Co-HITS, as well as a feature-rich regression model based on textural similarity. The results are conservative in all metrics (p@1, MAP, and nDCG) and in both introducing and non-introducing groups, showing that our ranking algorithm is potentially applicable to other tasks.

To sum up, the experimental results show that both our entity-based content introducing and \name\ are effective. When a stalemate occurs, the full \textsc{StalemateBreaker} yields a $+14.4\%$ boost of accuracy (p@1) compared with a state-of-the-practice system which our model is built upon.

\subsection{Analysis and Discussion}\label{ss:Analysis}
\begin{figure}[t]
\centering
\includegraphics[width=.48\textwidth]{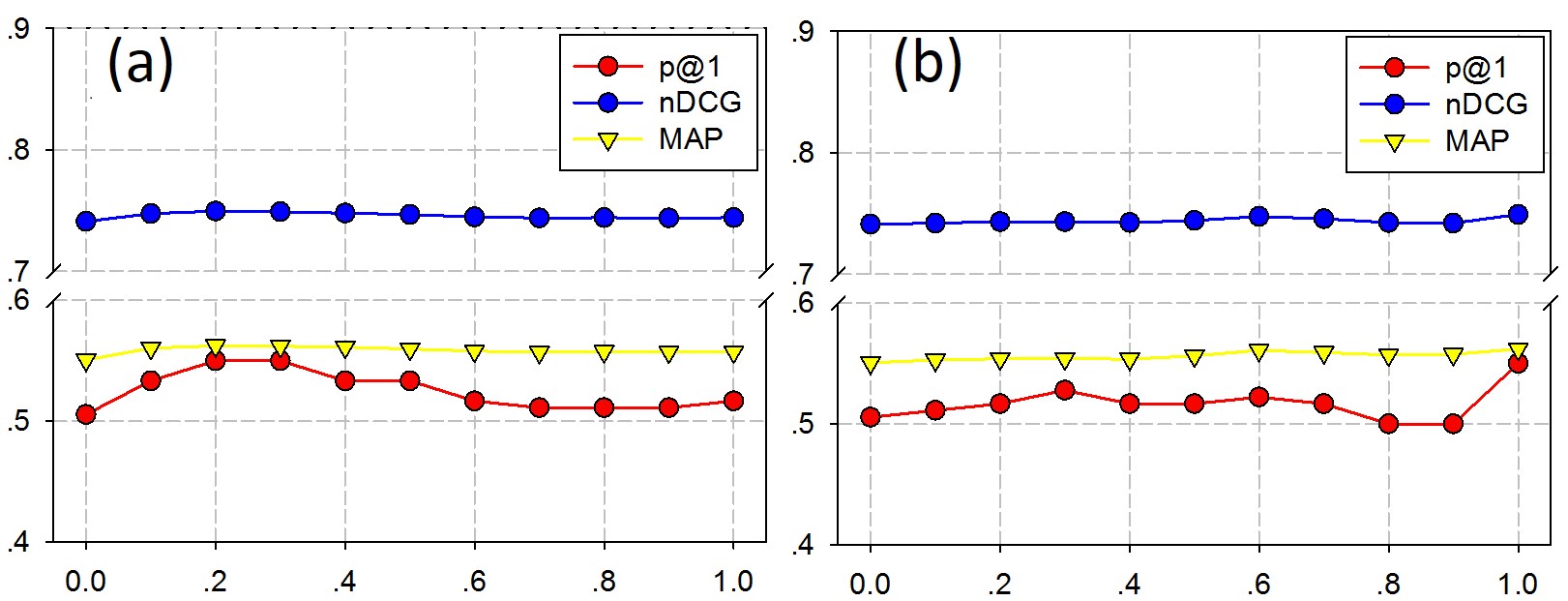}
\caption{Parameter analysis. (a) $\alpha_x$ and (b) $\alpha_y$ were tuned from $0$ to $1$ with a granulairty of $0.1$ given the other parameter was fixed (at the stationary point of grid search in the parameter space).
}
\label{fig:parameter}
\end{figure}
\begin{figure}[!t]
\centering
\includegraphics[width=.35\textwidth]{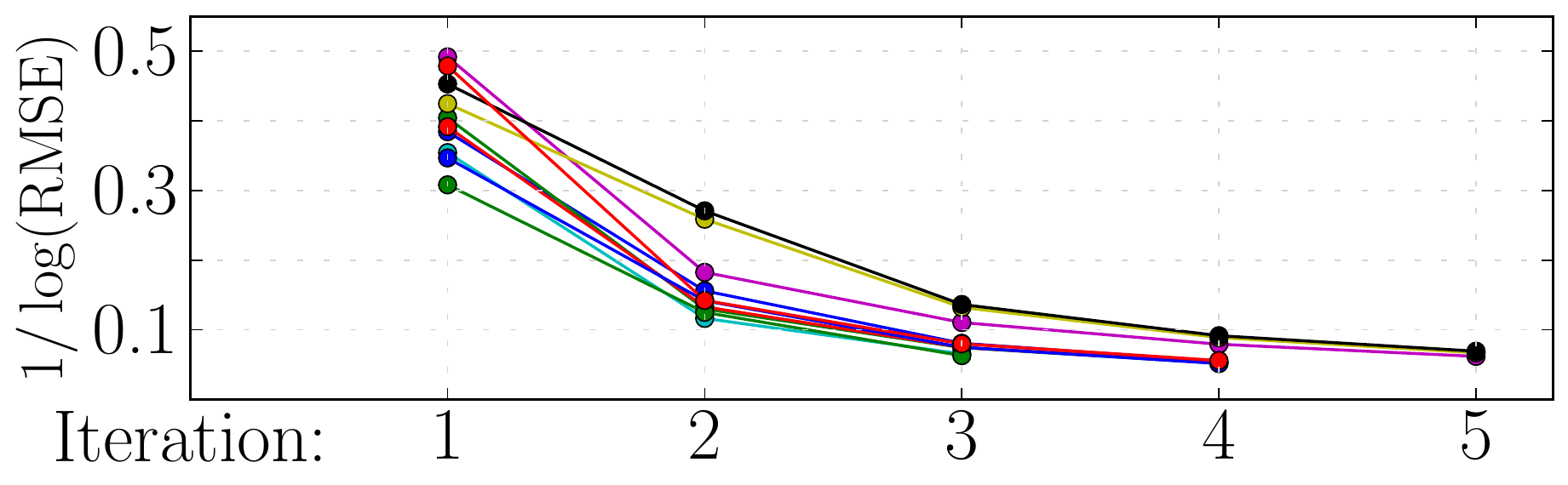}
\caption{Convergence analysis of global iterations of \name. 10 randomly chosen samples are plotted.}
\label{fig:convergence}
\end{figure}

\noindent\textbf{Parameter Settings.} In our \name\ model, we have three main parameters, $\mu$ in the PageRank phase, and $\alpha_x, \alpha_y$ in the HITS phase. $\mu$ was set to 0.15 following \citeauthor{a23}~\shortcite{a23} and not tuned in our experiment. For $\alpha_x$ and $\alpha_y$, we tried different values with a granularity of 0.1. The results are shown in Figure~\ref{fig:parameter}. If $\alpha$ is set to 0, the HITS update vanishes (the first term in Equations~\ref{eqn:HITS2}--\ref{eqn:HITS1}), and the system solely depends on ``prior'' information. The result is worse than using HITS update. 

When $\alpha$ increases, we observe an interesting phenomenon: queries and replies respond differently. For queries, i.e., context utterances, the performance peaks when $\alpha_x$ is small~(Figure~\ref{fig:parameter}a). This suggests that textual information $\hat{\mathbf x}$ (based on query-reply similarity) does recommend important queries. On the contrary, textual information $\hat{\mathbf y}$ for replies is inimical: the highest performance was achieved when $\alpha_y=1$, where $\hat{\mathbf y}$ was discarded. One plausible explanation is that, based on textual similarity, the algorithm is likely to recommend similar replies to the queries and be stuck in stalemate. In summary, the parameters we used were $\alpha_x=0.3$ and $\alpha_y=1$.


\smallskip
\noindent\textbf{Convergence.} In our \name, the convergence of each local step (either PageRank or HITS) is guaranteed because we have properly normalized the transition matrices. We further empirically analyze the global convergence by plotting the root mean square error between two successive HITS scores. Figure~\ref{fig:convergence} shows the curves of 10 randomly chosen samples; the results indicate that \name\ typically converges quickly in 3--5 global iterations. It works well in practice.

\smallskip
\noindent\textbf{Case Study.}
We illustrate two examples in Table~\ref{tab:case} obtained by \name. As seen, the (human) user doen not know waht to say by ``Err\dots,'' but our system can introduce new content (entities like ``WALL-E'' and the TW play ``Marriage of the Fairy Princess'') to attract users.

\begin{table}[!t]
\centering
\resizebox{.45\textwidth}{!}{
\begin{tabular}{rl}
\hline
Human:& 以后叫你伊娃。(I will call you Eva.)\\
Computer:& 伊娃不是那个机器人动画片？\\
& (Isn't Eva a robot cartoon?)\\
Human:& 你会聊天吗？(Can you talk?)\\
Computer:& 不是聊的好好的吗？\\
         & (Aren't we talking now?)\\
Human: & 啊\dots (Errr\dots)\\
Computer: & 我不是瓦力，我没有伊娃(I'm not\\
          & WALL-E and I don't have Eva.)\\
\multicolumn{2}{l}{\textbf{\emph{Remark: WALL-E and Eva are two robots in the movie WALL-E.}}}\\
\hline
Human: & 林依晨 (Yichen Lin.)\\
Computer: & 北川景子(Kitagawa Keiko.)\\
Human:& 我看电影去了(I will go to watch TV.)\\
Computer: & 不如看电影(Better go to a movie.)\\
Human: & 啊\dots(Errr\dots)\\
Computer: &我比较喜欢胡哥和林依晨的天外飞仙\\
          & (I like ``Marriage of the Fairy Princess''\\
          & casted by Ge Hu and Yichen Lin)\\
\multicolumn{2}{l}{\textbf{\emph{Remark: ``Marriage of the Fairy Princess'' is a TV play; Kitagawa} }}\\
\multicolumn{2}{l}{\textbf{\emph{Keiko, Yichen Lin, and Ge Hu are three actors/actresses.}}}\\
\hline
\end{tabular}
}
\caption{Examples obtained by \textsc{StalemateBreaker}.}
\label{tab:case}
\end{table}

\section{Conclusion}

In this paper, we addressed the problem of content introducing for stalemate breaking in open-domain conversation systems. We proposed a pipeline of content introducing based on an external knowledge graph. To enhance the interaction between queries (utterances in the conversation) and candidate replies (retrieved from a massive database), we further proposed the random walk-style \name\ reranking approach. Experiments show the effectiveness of both our content-introducing method and the ranking algorithm. 

\section*{Acknowledgments}

This paper is partially supported by the National Natural Science Foundation of China (Grant Nos.~61272343 and 61472006), the Doctoral Program of Higher Education of China (Grant No.~20130001110032), and the National Basic Research Program (973 Program No.~2014CB340405 and No.~2014CB340505).


\bibliographystyle{named}
\small
\bibliography{stalemate}

\end{CJK*}
\end{document}